\pdfoutput=1 

\documentclass[english]{gretsi}

\usepackage[utf8]{inputenc}

\usepackage[T1]{fontenc}

\usepackage{amsmath, amssymb, amsfonts}

\usepackage{caption}
\usepackage{xargs}
\usepackage{xurl}
\usepackage{xcolor}
\usepackage[mathscr]{euscript}
\usepackage{amsmath, amsfonts, amssymb, amsthm}
\usepackage{mathtools}
\usepackage{scalerel}
\usepackage[inline]{enumitem}
\usepackage{bm}

\usepackage{tikz, pgfplots}
\pgfplotsset{compat=1.18}
\usepgfplotslibrary{groupplots, fillbetween}

\allowdisplaybreaks

\let\originalleft\left
\let\originalright\right
\renewcommand{\left}{\mathopen{}\mathclose\bgroup\originalleft}
\renewcommand{\right}{\aftergroup\egroup\originalright}

\newtheorem{heuristic}{Heuristic}
\newtheorem{lemma}{Lemma}
\newtheorem{proposition}{Proposition}
\newtheorem{theorem}{Theorem}

\theoremstyle{definition}
\newtheorem{definition}{Definition}

\theoremstyle{remark}
\newtheorem{remark}{Remark}

\newcommandx{\lito}[2][1={}, 2={}]{\operatorname*{o_\mathnormal{#1}^\mathnormal{#2}}}
\newcommandx{\bigO}[2][1={}, 2={}]{\operatorname*{\mathcal{O}_\mathnormal{#1}^\mathnormal{#2}}}
\newcommandx{\bigTh}[2][1={}, 2={}]{\operatorname*{\Theta_\mathnormal{#1}^\mathnormal{#2}}}
\DeclareMathOperator*{\argmin}{\arg\min}
\DeclareMathOperator*{\argmax}{\arg\max}

\DeclareMathOperator{\Sp}{Sp}

\newcommand{\abs}[1]{\lvert #1 \rvert}
\newcommand{\Abs}[1]{\left\lvert #1 \right\rvert}
\newcommand{\deriv}[2]{\frac{\partial #1}{\partial #2}}
\newcommand{\esp}[1]{\mathbb{E}[#1]}
\newcommand{\Esp}[1]{\mathbb{E} \left[ #1 \right]}
\newcommand{\norm}[1]{\lVert #1 \rVert}
\newcommand{\Norm}[1]{\left\lVert #1 \right\rVert}
\newcommand{\scal}[2]{\langle #1, #2 \rangle}
\newcommand{\Scal}[2]{\left\langle #1, #2 \right\rangle}
\newcommand{\set}[1]{\{ #1 \}}

\newcommand{\bzero}{{\mathbf{0}}}

\newcommand{\ba}{{\bm{a}}}
\newcommand{\bb}{{\bm{b}}}
\newcommand{\bc}{{\bm{c}}}

\newcommand{\be}{{\bm{e}}}

\newcommand{\bu}{{\bm{u}}}
\newcommand{\bv}{{\bm{v}}}
\newcommand{\bw}{{\bm{w}}}
\newcommand{\bx}{{\bm{x}}}
\newcommand{\by}{{\bm{y}}}
\newcommand{\bz}{{\bm{z}}}

\newcommand{\bE}{{\bm{E}}}

\newcommand{\bI}{{\bm{I}}}

\newcommand{\bM}{{\bm{M}}}

\newcommand{\bQ}{{\bm{Q}}}

\newcommand{\bS}{{\bm{S}}}

\newcommand{\bV}{{\bm{V}}}

\newcommand{\bPhi}{{\bm{\Phi}}}

\newcommand{\scrA}{{\mathscr{A}}}
\newcommand{\scrB}{{\mathscr{B}}}

\newcommand{\scrN}{{\mathscr{N}}}

\newcommand{\scrT}{{\mathscr{T}}}

\newcommand{\bscrA}{{\bm{\mathscr{A}}}}
\newcommand{\bscrB}{{\bm{\mathscr{B}}}}

\newcommand{\bscrN}{{\bm{\mathscr{N}}}}

\newcommand{\bscrT}{{\bm{\mathscr{T}}}}

\newcommand{\calL}{{\mathcal{L}}}

\newcommand{\calN}{{\mathcal{N}}}

\newcommand{\bbC}{{\mathbb{C}}}

\newcommand{\bbR}{{\mathbb{R}}}

\newcommand{\rmd}{{\mathrm{d}}}

\newcommand{\rmi}{{\mathrm{i}}}

\newcommand{\rmF}{{\mathrm{F}}}

\definecolor{C0}{HTML}{1F77B4}
\definecolor{C1}{HTML}{FF7F0E}
\definecolor{C2}{HTML}{2CA02C}
\definecolor{C3}{HTML}{D62728}
\definecolor{C4}{HTML}{9467BD}
\definecolor{C5}{HTML}{8C564B}
\definecolor{C6}{HTML}{E377C2}
\definecolor{C7}{HTML}{7F7F7F}
\definecolor{C8}{HTML}{BCBD22}
\definecolor{C9}{HTML}{17BECF}

\begin{document}

\titre{Performance of Rank-One Tensor Approximation on Incomplete Data}

\auteurs{
  \auteur{Hugo}{Lebeau}{hugo.lebeau@inria.fr}{}
}

\affils{
  \affil{}{Inria, ENS de Lyon, CNRS, Université Claude Bernard Lyon 1, LIP, UMR 5668, 69342, Lyon cedex 07, France \\ \textit{work done while at} Université Grenoble Alpes, CNRS, Inria, Grenoble INP, LIG, UMR 5217, 38058, Grenoble cedex 09, France}
}


\resume{On s'intéresse à l'estimation d'un signal tensoriel de rang 1 lorsque seulement une portion $\varepsilon$ de son observation bruitée est disponible. Nous montrons que l'étude de ce problème se ramène à celle d'un modèle de matrice aléatoire dont l'analyse spectrale donne accès aux performances de reconstruction. Ces résultats mettent en lumière et précisent la perte de performance induite par une réduction artificielle du coût en mémoire d'un tenseur via la suppression d'une partie aléatoire de ses entrées.}

\abstract{We are interested in the estimation of a rank-one tensor signal when only a portion $\varepsilon$ of its noisy observation is available. We show that the study of this problem can be reduced to that of a random matrix model whose spectral analysis gives access to the reconstruction performance. These results shed light on and specify the loss of performance induced by an artificial reduction of the memory cost of a tensor via the deletion of a random part of its entries.}

\maketitle

\section{Introduction}

\noindent Processing large amounts of multimodal data is a common challenge of the big data era. Multiway arrays (tensors) have inherently a high volume, scaling exponentially with the number of modes, which can rapidly reach a memory limit. Often, the information sought in such gigantic raw data is a latent and low-dimensional structure. Hence, some techniques have been developed to compute tensor decompositions (such as the Canonical Polyadic Decomposition or the Multilinear SVD \cite{sidiropoulos_tensor_2017}) from \emph{incomplete} tensors \cite{vervliet_breaking_2014}. At first, they were used to tackle the issue of missing values, which is common in chemometrics data. Nowadays, they can be used to compute tensor decompositions on intentionally punctured data. That is, in order to reduce the volume of a tensor, only a small portion of its entries is actually stored and inference is performed on the incomplete tensor.

A simple yet effective procedure to reduce the volume of a huge tensor $\bscrT \in \bbR^{n_1 \times \dots \times n_d}$ is therefore to store a portion $\varepsilon \in [0, 1]$ of its entries chosen at random. Concretely, this is modeled by the entrywise multiplication $\bscrT \odot \bscrB$ where $\bscrB$ has i.i.d.\ $\text{Bernoulli}(\varepsilon)$ entries (i.e., $1$ with probability $\varepsilon$ and $0$ otherwise). It is then natural to ask: what is the loss of performance induced by this procedure? In other words, how different is the inference performed with $\bscrT \odot \bscrB$ from the one performed with $\bscrT$? A similar procedure on kernel matrices is studied using random matrix tools in \cite{couillet_two-way_2021}. It is shown therein that, when the dimension of the data is not too large, the number of entries of the kernel matrix can be drastically reduced with almost no impairment in the resulting estimation. Yet, in our case, puncturing $\odot \bscrB$ is directly applied onto the raw data $\bscrT$, which may cause a considerable performance loss. Our goal is therefore to quantify the latter in order to help deciding on the best trade-off between performance and memory cost. Our study also provides insights into the performance of algorithms relying on a similar procedure such as the CP-WOPT algorithm \cite{acar_scalable_2011, vervliet_breaking_2014} which computes a canonical polyadic decomposition from a tensor with missing data.

For the sake of simplicity we consider order-$3$ tensors $\bscrT \in \bbR^{n_1 \times n_2 \times n_3}$, the generalization to order-$d$ tensors $\bscrT \in \bbR^{n_1 \times \dots \times n_d}$ is straightforward. In order to study the estimation of a planted signal in $\bscrT$ from $\bscrT \odot \bscrB$, we consider the rank-one spiked tensor model $\bscrT = \beta \bx \otimes \by \otimes \bz + \frac{1}{\sqrt{N}} \bscrN$ where $\beta > 0$ is a parameter controlling the signal-to-noise ratio, $\bx, \by, \bz$ are fixed unit-norm vectors, $\otimes$ is the outer product ($\bx \otimes \by = \bx \by^\top$), $N = n_1 + n_2 + n_3$ and $\bscrN$ is an $n_1 \times n_2 \times n_3$ array with i.i.d.\ $\calN(0, 1)$ entries. We place ourselves in the large-dimensional regime where $n_1, n_2, n_3 \to +\infty$ with the ratios $(\frac{n_1}{N}, \frac{n_2}{N}, \frac{n_3}{N}) = (c_1, c_2, c_3)$ kept constant (and non-zero), simply denoted ``$N \to +\infty$''. This regime models the fact that all dimensions of the tensor are large, although they remain finite in practice. In this setting, we consider the estimation of $\beta, \bx, \by, \bz$ with the best rank-one approximation of $\bscrT \odot \bscrB$,
\begin{equation} \label{eq:estimation_problem}
(\sigma^\star, \bu^\star, \bv^\star, \bw^\star) \in \argmin_{\sigma, \bu, \bv, \bw} \Norm{\bscrT \odot \bscrB - \sigma \bu \otimes \bv \otimes \bw}_\rmF^2
\end{equation}
where the minimum is taken over $\sigma > 0$ and $\bu, \bv, \bw$ of unit norm. The ``vanilla'' rank-one estimation problem (i.e., without puncturing $\odot \bscrB$) is studied in \cite{seddik_when_2022}, where the analysis of its critical points is reduced to that of an associated random matrix model $\bPhi$ defined with contractions of the tensor on its singular vectors. This new approach allows the study of the rank-one approximation through that of a random matrix model, thereby allowing the harnessing of random matrix theory. Here, we follow the same approach to study the problem of missing data in the rank-one tensor estimation problem.

Yet, the entrywise multiplication $\odot \bscrB$ raises some mathematical difficulties and the rigorous analysis of this model is delicate. Therefore, we must rely on the following heuristic.

\begin{heuristic} \label{heu:independence}
A critical point $(\sigma^\star, \bu^\star, \bv^\star, \bw^\star)$ of Problem \eqref{eq:estimation_problem} is asymptotically (as $N \to +\infty$) independent of $\bscrB$.
\end{heuristic}

This property is hard to prove, but allows to write $\esp{\sum_{i, j, k} \scrT_{i, j, k} \scrB_{i, j, k} u^\star_i v^\star_j w^\star_k} = \varepsilon \esp{\sum_{i, j, k} \scrT_{i, j, k} u^\star_i v^\star_j w^\star_k} + \lito(1)$ and likewise for other similar ``contractions'' where the coefficient $\scrB_{i, j, k}$ can be replaced by $\varepsilon$, up to a vanishing term in the large $N$ limit. Furthermore, this heuristic leads to precise predictions which are verified by simulations (see Figure \ref{fig:eigvals} and Figure \ref{fig:reconstruction_epsilon}). The scope of this paper focuses on these results and their practical implications rather than the technical mathematical aspects implied by the entrywise multiplication $\odot \bscrB$, although they raise many interesting questions about the current state of our tools for tackling this kind of problems.

\paragraph{Notations.} The set $\set{1, \ldots, n}$ of positive integers smaller or equal to $n$ is denoted $[n]$. The canonical basis in $\bbR^n$ is $(\be^{(n)}_i)_{i \in [n]}$. Given a matrix $\bM \in \bbR^{n \times n}$, $\Sp \bM$ is the set of its eigenvalues. Given a tensor $\bscrA \in \bbR^{n_1 \times n_2 \times n_3}$ and three vectors $(\ba, \bb, \bc) \in \bbR^{n_1} \times \bbR^{n_2} \times \bbR^{n_3}$, the \emph{contraction} of $\bscrA$ onto $\ba, \bb, \bc$ is $\bscrA(\ba, \bb, \bc) = \sum_{i = 1}^{n_1} \sum_{j = 2}^{n_2} \sum_{k = 1}^{n_3} \scrA_{i, j, k} a_i b_j c_k$.

\section{Associated Random Matrix Model}

\subsection{First-Order Optimality Conditions}

\noindent Problem \eqref{eq:estimation_problem} can be equivalently restated as
\begin{equation} \label{eq:equivalent_problem}
(\bu^\star, \bv^\star, \bw^\star) \in \argmax_{\bu, \bv, \bw} \Abs{[\bscrT \odot \bscrB](\bu, \bv, \bw)}
\end{equation}
with $\sigma^\star = \abs{[\bscrT \odot \bscrB](\bu^\star, \bv^\star, \bw^\star)}$. The Lagrangian associated to this problem is
\[
\calL = [\bscrT \odot \bscrB](\bu, \bv, \bw) - \frac{1}{2} \begin{bmatrix} \lambda_u \\ \lambda_v \\ \lambda_w \end{bmatrix} \cdot \begin{bmatrix} \Norm{\bu}^2 - 1 \\ \Norm{\bv}^2 - 1 \\ \Norm{\bw}^2 - 1 \end{bmatrix}
\]
where $\lambda_u, \lambda_v, \lambda_w$ are Lagrange multipliers associated to the unit-norm constraints and $\cdot$ denotes the dot product. From the first-order optimality conditions $\nabla \calL = \bzero$, we find that $\sigma^\star, \bu^\star, \bv^\star, \bw^\star$ must satisfy
\begin{equation} \label{eq:CO1}
\left\{\begin{array}{l}
{[}\bscrT \odot \bscrB](:, \bv^\star, \bw^\star) = \sigma^\star \bu^\star \\
{[}\bscrT \odot \bscrB](\bu^\star, :, \bw^\star) = \sigma^\star \bv^\star \\
{[}\bscrT \odot \bscrB](\bu^\star, \bv^\star, :) = \sigma^\star \bw^\star
\end{array}\right. .
\end{equation}

\subsection{Critical Points}
\label{sec:associated_matrix:critical_points}

\noindent As our goal is to reconstruct the signal $\beta \bx \otimes \by \otimes \bz$ from the observation $\bscrT \odot \bscrB$, we seek the behavior of $\sigma^\star$ (as an estimator of $\beta$) and the alignments of $\bu^\star, \bv^\star, \bw^\star$ with $\bx, \by, \bz$. These quantities are expected to \emph{concentrate} around their mean, therefore it is sufficient to study their expectations. To that end, we have the following ``Gaussian integration by parts'' lemma.

\begin{lemma}[Stein]
Let $Z \sim \calN(0, 1)$ and $f : \bbR \to \bbC$ be a polynomially bounded differentiable function such that $\esp{\abs{f'(Z)}} < +\infty$. Then, $\esp{Z f(Z)} = \esp{f'(Z)}$.
\end{lemma}

Based on this lemma and Heuristic \ref{heu:independence}, we can derive the following relations
\small
\begin{gather*}
\Esp{\sigma^\star} \asymp \varepsilon \beta \Esp{\Scal{\bx}{\bu^\star} \Scal{\by}{\bv^\star} \Scal{\bz}{\bw^\star}} + \frac{\varepsilon}{\sqrt{N}} \sum_{i, j, k} \Esp{\deriv{u^\star_i v^\star_j w^\star_k}{\scrN_{i, j, k}}} \\
\Esp{\sigma^\star \Scal{\bx}{\bu^\star}} \asymp \varepsilon \beta \Esp{\Scal{\by}{\bv^\star} \Scal{\bz}{\bw^\star}} + \frac{\varepsilon}{\sqrt{N}} \sum_{i, j, k} x_i \Esp{\deriv{v^\star_j w^\star_k}{\scrN_{i, j, k}}} \\
\Esp{\sigma^\star \Scal{\by}{\bv^\star}} \asymp \varepsilon \beta \Esp{\Scal{\bx}{\bu^\star} \Scal{\bz}{\bw^\star}} + \frac{\varepsilon}{\sqrt{N}} \sum_{i, j, k} y_j \Esp{\deriv{u^\star_i w^\star_k}{\scrN_{i, j, k}}} \\
\Esp{\sigma^\star \Scal{\bz}{\bw^\star}} \asymp \varepsilon \beta \Esp{\Scal{\bx}{\bu^\star} \Scal{\by}{\bv^\star}} + \frac{\varepsilon}{\sqrt{N}} \sum_{i, j, k} z_k \Esp{\deriv{u^\star_i v^\star_j}{\scrN_{i, j, k}}}
\end{gather*}
\normalsize
where $\asymp$ means equality up to a vanishing term as $N \to +\infty$.

\begin{remark}
In order to avoid any differentiability issues, we consider, from now on, $(\sigma^\star, \bu^\star, \bv^\star, \bw^\star)$ as a fixed \emph{critical point} of Problem \eqref{eq:estimation_problem} rather than a global minimizer.
\end{remark}

\begin{remark}
The concentration of relevant quantities stems from standard arguments such as the Poincaré-Nash inequality for Gaussian vectors (see Lemma 2.14 in \cite{couillet_random_2022}) and the sub\-gaussianity of the entries of $\bscrN \odot \bscrB$.
\end{remark}

From the previous relations, we see that it is necessary to find an expression for the derivatives of $\bu^\star, \bv^\star, \bw^\star$ with respect to the entries of $\bscrN$. Despite their intricate dependence on the noise, differentiating conditions given in Equation \eqref{eq:CO1} leads to the following result.

\begin{proposition} \label{prop:deriv_u}
Define the $N \times N$ matrix
\[
\bPhi(\bu^\star, \bv^\star, \bw^\star) = \begin{bmatrix}
\bzero_{n_1 \times n_1} & \bPhi^{(1, 2)} & \bPhi^{(1, 3)} \\
\bPhi^{(1, 2) \top} & \bzero_{n_2 \times n_2} & \bPhi^{(2, 3)} \\
\bPhi^{(1, 3) \top} & \bPhi^{(2, 3) \top} & \bzero_{n_3 \times n_3}
\end{bmatrix}
\]
with $\bPhi^{(1, 2)} = [\bscrT \odot \bscrB](:, :, \bw^\star)$, $\bPhi^{(1, 3)} = [\bscrT \odot \bscrB](:, \bv^\star, :)$ and $\bPhi^{(2, 3)} = [\bscrT \odot \bscrB](\bu^\star, :, :)$. If $\sigma^\star$ is \emph{not} an eigenvalue of $\bPhi(\bu^\star, \bv^\star, \bw^\star)$, then, for all $(i, j, k) \in [n_1] \times [n_2] \times [n_3]$,
\[
\deriv{}{\scrN_{i, j, k}} \begin{bmatrix}
\bu^\star \\ \bv^\star \\ \bw^\star
\end{bmatrix}
= -\frac{\scrB_{i, j, k}}{\sqrt{N}} \bQ(\sigma^\star) \begin{bmatrix}
v^\star_j w^\star_k (\be^{(n_1)}_i - u^\star_i \bu^\star) \\
u^\star_i w^\star_k (\be^{(n_2)}_j - v^\star_j \bv^\star) \\
u^\star_i v^\star_j (\be^{(n_3)}_k - w^\star_k \bw^\star)
\end{bmatrix}
\]
where $\bQ(\sigma^\star) = [\bPhi(\bu^\star, \bv^\star, \bw^\star) - \sigma_\star \bI_N]^{-1}$.
\end{proposition}

This Proposition shows the link between the derivatives of $\bu^\star, \bv^\star, \bw^\star$ and the associated matrix model $\bPhi(\bu^\star, \bv^\star, \bw^\star)$ through a matrix $\bQ(\sigma^\star)$, known as the \emph{resolvent} --- a common tool in random matrix theory which is used to study the spectral properties of random matrices \cite{couillet_random_2022}. The reconstruction performance of $(\sigma^\star, \bu^\star, \bv^\star, \bw^\star)$ can therefore be revealed by a spectral analysis of $\bPhi(\bu^\star, \bv^\star, \bw^\star)$.

\subsection{First Remarks on the Associated Matrix}

\noindent Notice that $\bPhi(\bu^\star, \bv^\star, \bw^\star)$ can be decomposed as
\begin{equation} \label{eq:decomposition_spike}
\bPhi(\bu^\star, \bv^\star, \bw^\star) = \varepsilon \beta \bV \bS \bV^\top + \bPhi_0(\bu^\star, \bv^\star, \bw^\star) + \bE
\end{equation}
where $\bV = \begin{bsmallmatrix} \bx & \bzero_{n_1} & \bzero_{n_1} \\ \bzero_{n_2} & \by & \bzero_{n_2} \\ \bzero_{n_3} & \bzero_{n_3} & \bz \end{bsmallmatrix}$, $\bS = \begin{bsmallmatrix} 0 & \scal{\bz}{\bw^\star} & \scal{\by}{\bv^\star} \\ \scal{\bz}{\bw^\star} & 0 & \scal{\bx}{\bu^\star} \\ \scal{\by}{\bv^\star} & \scal{\bx}{\bu^\star} & 0 \end{bsmallmatrix}$, $\bPhi_0$ is defined as $\bPhi$ but with $\bscrT$ replaced by $\frac{1}{\sqrt{N}} \bscrN$ (i.e., $\beta = 0$) and $\norm{\bE} \to 0$ almost surely as $N \to +\infty$ (Heuristic \ref{heu:independence}). Hence, $\bPhi(\bu^\star, \bv^\star, \bw^\star)$ has the structure of a \emph{spiked model} --- it is a rank-$3$ perturbation of a random matrix. Moreover, based on the definition of $\bPhi(\bu^\star, \bv^\star, \bw^\star)$ and a second-order optimality condition of Problem \eqref{eq:equivalent_problem}, we can state the following properties on its spectrum.

\begin{proposition}
\begin{enumerate}
\item $2 \sigma^\star$ is an eigenvalue of $\bPhi(\bu^\star, \bv^\star, \bw^\star)$ with multiplicity $1$ and eigenvector $\begin{bsmallmatrix} \bu^\star \\ \bv^\star \\ \bw^\star \end{bsmallmatrix}$.
\item $-\sigma^\star$ is an eigenvalue of $\bPhi(\bu^\star, \bv^\star, \bw^\star)$ with multiplicity (at least) $2$ and eigenspace spanned by $\begin{bsmallmatrix} \bu^\star & \bu^\star \\ -\bv^\star & \bzero_{n_2} \\ \bzero_{n_3} & -\bw^\star \end{bsmallmatrix}$.
\item For all eigenvalue $\lambda$ of $\bPhi(\bu^\star, \bv^\star, \bw^\star)$, either $\lambda = 2 \sigma^\star$ or $\abs{\lambda} \leqslant \sigma^\star$.
\end{enumerate}
\end{proposition}

These properties are illustrated by the histograms of eigenvalues depicted in Figure \ref{fig:eigvals}.

\section{Random Matrix Analysis}

\subsection{Limiting Spectral Distribution}

\noindent In order to study the eigenvalue distribution of the associated random matrix $\bPhi$, we must introduce an important tool, namely, the Stieltjes transform of a probability distribution.

\begin{definition}
Given a real probability measure $\mu$, its \emph{Stieltjes transform} is $m_\mu : z \in \bbC \setminus \bbR \mapsto \int_\bbR \frac{\rmd \mu(t)}{t - z}$.
\end{definition}

The Stieltjes transform satisfies many interesting properties (see \cite{couillet_random_2022}) and the underlying probability measure can be reconstructed with the inverse Stieltjes transform formula $\mu([a, b]) = \frac{1}{\pi} \lim_{\eta \downarrow 0} \int_a^b \Im[m_\mu(x + \rmi \eta)] \rmd x$ provided that $a < b$ are not atoms of $\mu$.

Firstly, we study the limiting behavior of the \emph{empirical spectral distribution} $\mu_N = \frac{1}{N} \sum_{\lambda \in \Sp \bPhi(\bu^\star, \bv^\star, \bw^\star)} \delta_\lambda$. From the decomposition \eqref{eq:decomposition_spike}, we can already expect that it has the same behavior as that of $\bPhi_0(\bu^\star, \bv^\star, \bw^\star)$ since the rank-$3$ perturbation will have a negligible weight. Yet, the entries of $\bPhi_0(\bu^\star, \bv^\star, \bw^\star)$ show non-trivial dependencies because of the contractions on $\bu^\star, \bv^\star, \bw^\star$. Nevertheless, these dependencies vanish asymptotically as precised in the following theorem.

\begin{theorem} \label{thm:lsd}
Assume Heuristic \ref{heu:independence} is true. Let $\ba, \bb, \bc$ be unit-norm vectors in dimension $n_1, n_2, n_3$ respectively. The empirical spectral distributions of $\bPhi_0(\ba, \bb, \bc)$ and $\bPhi(\bu^\star, \bv^\star, \bw^\star)$ both converge weakly almost surely to the same probability distribution $\bar{\mu}$ whose Stieltjes transform is $\bar{m} = m_1 + m_2 + m_3$ with $m_\ell$ ($\ell \in \set{1, 2, 3}$) such that, for all $z \in \bbC \setminus \bbR$,\footnote{Recall that $c_\ell = n_\ell / N$.}
\begin{equation} \label{eq:fixed_point}
\varepsilon m_\ell(z) (\bar{m}(z) - m_\ell(z)) + z m_\ell(z) + c_\ell = 0.
\end{equation}
\end{theorem}

\begin{figure*}
\centering
\input{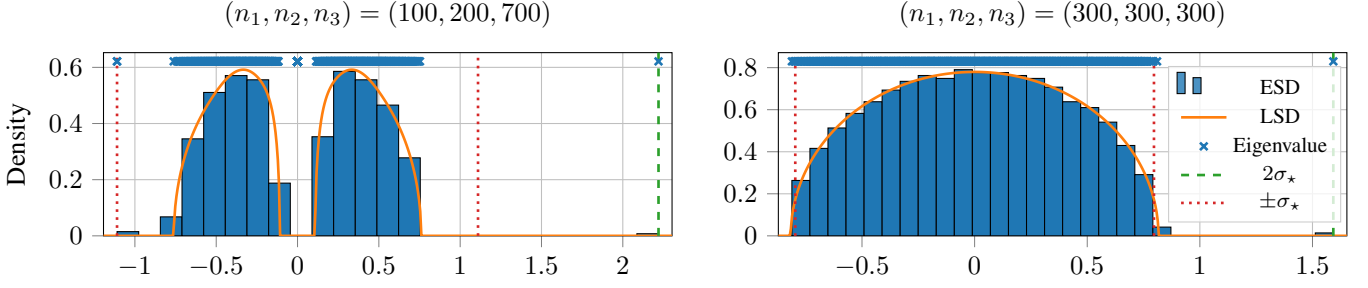}
\caption{Empirical spectral distribution (ESD) and limiting spectral distribution (ESD) of $\bPhi(\bu^\star, \bv^\star, \bw^\star)$ with two different tensor shapes. \textbf{Parameters}: $\beta = 4$, $\varepsilon = 0.25$. The histogram on the \textbf{left} only counts the $599$ (out of $1\,000$) \emph{non-zero} eigenvalues.}
\label{fig:eigvals}
\end{figure*}

The Stieltjes transform $\bar{m}$ of the limiting spectral distribution $\bar{\mu}$ is the sum of $3$ Stieltjes transforms $m_\ell$ which satisfy the fixed-point equation \eqref{eq:fixed_point}. In two different settings, Figure \ref{fig:eigvals} shows good agreement between the empirical spectral distribution of $\bPhi(\bu^\star, \bv^\star, \bw^\star)$ and the density of $\bar{\mu}$ predicted by Theorem \ref{thm:lsd}. If $n_1 = n_2 = n_3$, we see that $\bar{\mu}$ corresponds to a semicircle distribution. This observation can be justified by the fact that, in this case, $m_1 = m_2 = m_3$ therefore $\bar{m}$ satisfies
\[
\frac{2 \varepsilon}{3} \bar{m}^2(z) + z \bar{m}(z) + 1 = 0, \qquad z \in \bbC \setminus \bbR,
\]
which is the equation characterizing the Stieltjes transform of the semicircle distribution on $[-2 \sqrt{2 \varepsilon / 3}, +2 \sqrt{2 \varepsilon / 3}]$.

\begin{remark}[Universality]
With the change of variables $(z, m(z)) \curvearrowright (\sqrt{\varepsilon} \tilde{z}, \frac{1}{\sqrt{\varepsilon}} \tilde{m}(\tilde{z}))$ applied to Equation \eqref{eq:fixed_point}, we recover the relation given in \cite{seddik_when_2022} for the model without puncturing ($\varepsilon = 1$). Hence, the system of equations satisfied by $m_\ell(z)$ is the same as that of the model \emph{without} mask $\bscrB$ but with an $\calN(0, \varepsilon)$-noise instead of $\calN(0, 1)$. This suggests that the limiting spectral distribution of $\bPhi(\bu^\star, \bv^\star, \bw^\star)$ only depends on the first and second moments of the entries of $\bscrT$.
\end{remark}

\subsection{Asymptotic Spike Recovery}

\noindent Relying on our previous spectral analysis of $\bPhi(\bu^\star, \bv^\star, \bw^\star)$ and Proposition \ref{prop:deriv_u}, we can now express the derivatives of $\bu^\star, \bv^\star, \bw^\star$ with respect to the entries of $\bscrN$ and describe the asymptotic behavior of $\sigma^\star$ and the alignments $\scal{\bx}{\bu^\star}, \scal{\by}{\bv^\star}, \scal{\bz}{\bw^\star}$ thanks to the relations found with Stein's lemma in Section \ref{sec:associated_matrix:critical_points}.

\begin{theorem} \label{thm:spike}
Assume Heuristic \ref{heu:independence} is true. There exists a threshold $\beta_s > 0$ such that, as $N \to +\infty$, if $\beta > \beta_s$ then $\sigma^\star \to \sigma_\infty$ almost surely and
\[
\Abs{\Scal{\bx}{\bu^\star}} \xrightarrow{\text{a.s.}} q_1, \quad \Abs{\Scal{\by}{\bv^\star}} \xrightarrow{\text{a.s.}} q_2, \quad \Abs{\Scal{\bz}{\bw^\star}} \xrightarrow{\text{a.s.}} q_3,
\]
where, for $\ell \in \set{1, 2, 3}$, $q_\ell^2 = 1 - \frac{\varepsilon m_\ell^2(\sigma_\infty)}{c_\ell}$ and $\sigma_\infty$ satisfies
\begin{equation} \label{eq:fixed_point_sigma}
\sigma_\infty + \varepsilon \bar{m}(\sigma_\infty) - \varepsilon \beta q_1 q_2 q_3 = 0.
\end{equation}
\end{theorem}

Figure \ref{fig:sigma_align} depicts the values of $\sigma_\infty$ and $q_1, q_2, q_3$ against $\beta$ in the same settings as Figure \ref{fig:eigvals} (where $\beta$ was fixed to $4$). We observe a phase transition phenomenon: it is only above the threshold value $\beta_s$ that $\sigma_\infty$ is well-defined and $q_1, q_2, q_3 > 0$. As $\beta$ crosses the threshold, the alignments suddenly jump from $0$ to a high value, revealing the appearance of an informative critical point of Problem \eqref{eq:estimation_problem} as described in \cite{jagannath_statistical_2020}. Then, as $\beta$ increases, the alignments converge to $1$ while $\sigma_\infty$ approaches its asymptote $\varepsilon \beta$ from above. 

In the ``cubic setting'' $c_1 = c_2 = c_3$, we have the closed-form expression $\beta_s = \sqrt{\frac{d - 1}{\varepsilon d}} \left( \frac{d - 1}{d - 2} \right)^{\frac{d - 2}{2}}$ which we state for a generic tensor order $d \geqslant 3$ to show that it coincides with the value found in \cite{jagannath_statistical_2020} for a symmetric spike without puncturing. Moreover, we can also precise the value of the alignments right after the phase transition, $\lim_{\beta \downarrow \beta_s} q_\ell = \sqrt{\frac{d - 2}{d - 1}}$, which surprisingly does not depend $\varepsilon$. This shows that keeping less entries of $\bscrT$ (reducing $\varepsilon$) only causes the phase transition to recede ($\beta_s \propto \frac{1}{\sqrt{\varepsilon}}$) but does not change the alignments after $\beta_s$.

\begin{figure*}
\centering
\captionbox{
Asymptotic values of $\sigma^\star$ (\textbf{top}) and alignments $\abs{\scal{\bx}{\bu^\star}}, \abs{\scal{\by}{\bv^\star}}, \abs{\scal{\bz}{\bw^\star}}$ (\textbf{bottom}) predicted by Theorem \ref{thm:spike} against $\beta$ with $\varepsilon = 0.25$ and tensor shapes $(c_1, c_2, c_3)$ corresponding to the two settings of Figure \ref{fig:eigvals}.
\label{fig:sigma_align}
}{\input{figures/sigma_align}\vspace{-5pt}}
\hfill
\captionbox{
Empirical alignments and their asymptotic values against $\varepsilon$ with $\beta = 2.5$ and tensor shape $(\frac{1}{10}, \frac{2}{10}, \frac{7}{10})$. Dots correspond to mean alignments averaged over $20$ runs and the corresponding filled areas show the standard deviation.
\label{fig:reconstruction_epsilon}
}{\begin{tikzpicture}
\begin{axis}[
width=.45\linewidth,
height=.28\linewidth,
title={$(n_1, n_2, n_3) = (50, 100, 350)$},
legend style={at={(1-0.02*28/45, 0.02)}, anchor=south east, fill opacity=0.8, draw opacity=1, text opacity=1, draw=white!80!black, font=\footnotesize},
tick align=outside,
tick pos=left,
xmajorgrids,
xmin=0, xmax=1,
xlabel={$\varepsilon$},
ymajorgrids,
ymin=0, ymax=1,
ylabel={Alignment}
]
\addplot [line width=1pt, C0, only marks, mark size=1pt] table {figures/align_mean_C0.tab};
\addlegendentry{$\abs{\scal{\bx}{\bu^\star}}$}
\addplot [line width=1pt, C1, only marks, mark size=1pt] table {figures/align_mean_C1.tab};
\addlegendentry{$\abs{\scal{\by}{\bv^\star}}$}
\addplot [line width=1pt, C2, only marks, mark size=1pt] table {figures/align_mean_C2.tab};
\addlegendentry{$\abs{\scal{\bz}{\bw^\star}}$}
\addplot [name path=C0b, line width=1pt, C0, opacity=.2, forget plot] table {figures/align_mean-std_C0.tab};
\addplot [name path=C1b, line width=1pt, C1, opacity=.2, forget plot] table {figures/align_mean-std_C1.tab};
\addplot [name path=C2b, line width=1pt, C2, opacity=.2, forget plot] table {figures/align_mean-std_C2.tab};
\addplot [name path=C0t, line width=1pt, C0, opacity=.2, forget plot] table {figures/align_mean+std_C0.tab};
\addplot [name path=C1t, line width=1pt, C1, opacity=.2, forget plot] table {figures/align_mean+std_C1.tab};
\addplot [name path=C2t, line width=1pt, C2, opacity=.2, forget plot] table {figures/align_mean+std_C2.tab};
\addplot [C0, opacity=.2, forget plot] fill between [of=C0b and C0t];
\addplot [C1, opacity=.2, forget plot] fill between [of=C1b and C1t];
\addplot [C2, opacity=.2, forget plot] fill between [of=C2b and C2t];
\addplot [line width=1pt, C0, dashed, forget plot] table {
0 0
0.16 0
};
\addplot [line width=1pt, C0, dashed] table {figures/align_pred_C0.tab};
\addlegendentry{$q_1$}
\addplot [line width=1pt, C1, dashed, forget plot] table {
0 0
0.16 0
};
\addplot [line width=1pt, C1, dashed] table {figures/align_pred_C1.tab};
\addlegendentry{$q_2$}
\addplot [line width=1pt, C2, dashed, forget plot] table {
0 0
0.16 0
};
\addplot [line width=1pt, C2, dashed] table {figures/align_pred_C2.tab};
\addlegendentry{$q_3$}
\end{axis}
\end{tikzpicture}}
\end{figure*}

\subsection{Performance vs.\ Cost Trade-Off}

\noindent Given a punctured tensor $\bscrT \odot \bscrB$ with $\bscrT = \beta \bx \otimes \by \otimes \bz + \frac{1}{\sqrt{N}} \bscrN$, the reconstruction performance of $\sigma^\star, \bu^\star, \bv^\star, \bw^\star$ hinges upon the strength of the signal $\beta$ and the portion of non-missing values $\varepsilon$. Although $\beta$ is fixed by the problem, $\varepsilon$ may be chosen by the user in a context where the data is intentionally punctured to store the tensor at a lower memory cost.

From a general perspective, applying the change of variables $(z, m(z)) \curvearrowright (\sqrt{\varepsilon} \tilde{z}, \frac{1}{\sqrt{\varepsilon}} \tilde{m}(\tilde{z}))$ to Equation \eqref{eq:fixed_point_sigma} shows that the reconstruction performance of the best rank-one approximation of $\bscrT \odot \bscrB$ is the same as the performance achieved \emph{without} puncturing ($\varepsilon = 1$) but with a signal strength $\tilde{\beta} = \sqrt{\varepsilon} \beta$ instead. This dilation by a factor $\sqrt{\varepsilon}$ summarizes the impact of puncturing on the reconstruction performance. For small values of $\varepsilon$ this may represent a significant loss of performance.

Figure \ref{fig:reconstruction_epsilon} displays the empirical alignments computed on a $50 \times 100 \times 350$ tensor $\bscrT$ with $\beta = 2.5$ as a function of the puncturing level $\varepsilon$. Theoretical curves also reveal a phase transition at a threshold value of $\varepsilon$ (here, approximately $0.17$). Yet, it is only for values of $\varepsilon \gtrsim 0.53$ that empirical alignments match the predictions of Theorem \ref{thm:spike}. This reveals a \emph{computational-to-statistical gap}: $0.17 \lesssim \varepsilon \lesssim 0.53$ corresponds to a \emph{hard phase} where recovery is statistically possible but computationally too difficult. The problem becomes computationally feasible above the \emph{algorithmic threshold} which can also be characterized with a random matrix analysis, as in \cite{lebeau_random_2025}.

\section{Conclusion and Perspectives}

\noindent Studying the performances of inference on a punctured tensor $\bscrT \odot \bscrB$ is of interest in a context of missing data or memory limitations. Here, we study its best rank-one approximation and show that the behavior of the critical points of Problem \eqref{eq:estimation_problem} is linked to a random matrix $\bPhi$. The spectral analysis of the latter gives access to the performance of reconstruction of a planted rank-one signal $\beta \bx \otimes \by \otimes \bz$ in $\bscrT$. In particular, the reconstruction performance from the punctured tensor $\bscrT \odot \bscrB$ is the same as the performance achieved from $\bscrT$ directly but with a smaller signal strength $\tilde{\beta} = \sqrt{\varepsilon} \beta$ instead. Although Theorem \ref{thm:spike} describes the alignment with the signal of an informative optimum to Problem \eqref{eq:estimation_problem}, in practice, recovering this optimum is hard and becomes feasible only if the signal strength is above an \emph{algorithmic threshold}, whose theoretical behavior is also accessible with random matrix tools, see \cite{lebeau_random_2025}.

The analysis of punctured tensors is rich in exciting challenges. On the theoretical side, the rigorous justification of Heuristic \ref{heu:independence} requires tools to untangle the dependence of solutions to Problem \eqref{eq:estimation_problem} on the entries of $\bscrB$. Furthermore, in order to face strong memory limitations, more clever data-dependent puncturing approaches should be considered. Although their theoretical study is expected to be much more delicate than that of random puncturing (considered here), relevant policies such as \emph{sparsification} via hard thresholding or even \emph{quantization} of the entries may prove to be much more efficient.

{\small \bibliography{bibliography}}

\end{document}